\documentclass[letterpaper, 10 pt, conference]{ieeeconf}  

\IEEEoverridecommandlockouts                              

\usepackage{graphicx} 
\usepackage{amsmath} 
\usepackage{amssymb}  

\makeatletter
\let\MYcaption\@makecaption
\makeatother
\usepackage[font=footnotesize]{subcaption}
\makeatletter
\let\@makecaption\MYcaption
\makeatother

\usepackage{hyperref}

\usepackage{tabularx}
\usepackage{booktabs}
\usepackage{multirow}
\usepackage{makecell}
\usepackage{tikz}

\usepackage[noadjust]{cite}

\usepackage{gensymb}
\usepackage[gen]{eurosym}

\newcommand{\eg}{\textit{e.g.}}

\title{\LARGE \bf
RealAnt: An Open-Source Low-Cost Quadruped for Education and Research in Real-World Reinforcement Learning
}

\author{Rinu Boney$^{*1}$, Jussi Sainio$^{*2}$, Mikko Kaivola$^{1}$, Arno Solin$^{1}$, and Juho Kannala$^{1}$
\thanks{$^*$ Equal contribution}
\thanks{$^{1}$ Aalto University, Finland,
        {\tt\small firstname.lastname@aalto.fi}}%
\thanks{$^{2}$ Ote Robotics Ltd, Finland,
        {\tt\small firstname@oterobotics.com}}%
}

\begin{document}

\maketitle
\thispagestyle{empty}
\pagestyle{empty}

\begin{abstract}
Current robot platforms available for research are either very expensive or unable to handle the abuse of exploratory controls in reinforcement learning.
We develop RealAnt, a minimal low-cost physical version of the popular `Ant' benchmark used in reinforcement learning. RealAnt costs only $\sim$350~\euro{} (\$410) in materials and can be assembled in less than an hour. We validate the platform with reinforcement learning experiments and provide baseline results on a set of benchmark tasks. 
We demonstrate that the RealAnt robot can learn to walk from scratch from less than 10 minutes of experience.
We also provide simulator versions of the robot (with the same dimensions, state-action spaces, and delayed noisy observations) in the MuJoCo and PyBullet simulators.
We open-source hardware designs, supporting software, and baseline results for educational use and reproducible research.
\end{abstract}

\section{Introduction}

The field of reinforcement learning (RL) has advanced significantly in recent years, with numerous success stories in solving challenging control problems.
This is largely due to the availability of simulators that allow for rapid testing of algorithmic performance, which are inexpensive, fast, and can be run in parallel.
However, simulators often make unrealistic assumptions about the world. For example, the popular simulator benchmarks for RL \cite{brockman2016openai, tassa2018deepmind, coumans2019pybullet} present no communications delays or noise, have simple dynamics, allow for environment resets, and have no concerns about the safety or durability of the robot hardware \cite{dulac2019challenges}.
We have to bridge this gap by grounding the development of reinforcement learning algorithms on real-world problems such as robot learning.

Most research on robotics is conducted on industrial robots that are very expensive, costing thousands of dollars. 
This is not very affordable to all researchers, let alone educational use.
Traditional control algorithms require precise hardware that is easy to model. This places a lot of limitations on robot design.
Reinforcement learning algorithms are able to learn controllers without modeling the dynamics and can also handle noisy observations and controls. However, aggressive exploratory actions taken by RL algorithms can easily damage the components of a robot. For example, plastic gears in RC servos or naively designed 3D printed parts can easily break during random exploration and learning. Also, many educational robot platforms do not offer a 
pose estimation or tracking solution, which means one cannot utilize reinforcement learning or any closed-loop algorithms for control.
In this paper, we present a minimal low-cost quadruped robot platform that can support and sustain reinforcement learning.

We develop and validate RealAnt, a physical version of the popular {\em Ant benchmark} available in OpenAI Gym \cite{brockman2016openai}, DeepMind Control Suite \cite{tassa2018deepmind}, and PyBullet \cite{coumans2019pybullet} simulators. The Ant benchmark, introduced in \cite{schulman2016high}, involves learning a controller for an 8~DoF quadruped robot to walk forward as fast as possible. 
The Ant benchmark is very widely used in the RL community, allowing for a shallow learning curve in using the RealAnt robot introduced in this paper.

RealAnt is a quadruped robot and learning to walk requires delicate balancing and coordination of the leg joints.
Learning controllers using RL for such a quadruped robot is challenging.
While state-of-the-art RL algorithms are able to learn to walk in a reasonable amount of time \cite{haarnoja2018softb}, there is still room for improvements in sample-efficiency and safe exploration, which are active areas of research in RL.
The Ant benchmark strikes a good balance between simplicity and complexity in terms of the difficulty in building the robot and controlling it. The simple design of the 8~DoF quadruped allows us to build it in a cost-efficient and scalable manner.

\begin{figure}[t]
  \centering
  \begin{tikzpicture}[outer sep=0]
    \node[minimum width=\columnwidth,minimum height=0.71\columnwidth,rounded corners=2mm,path picture={
      \node at (path picture bounding box.center){
        \includegraphics[width=\linewidth, trim=0 0 0 200, clip]{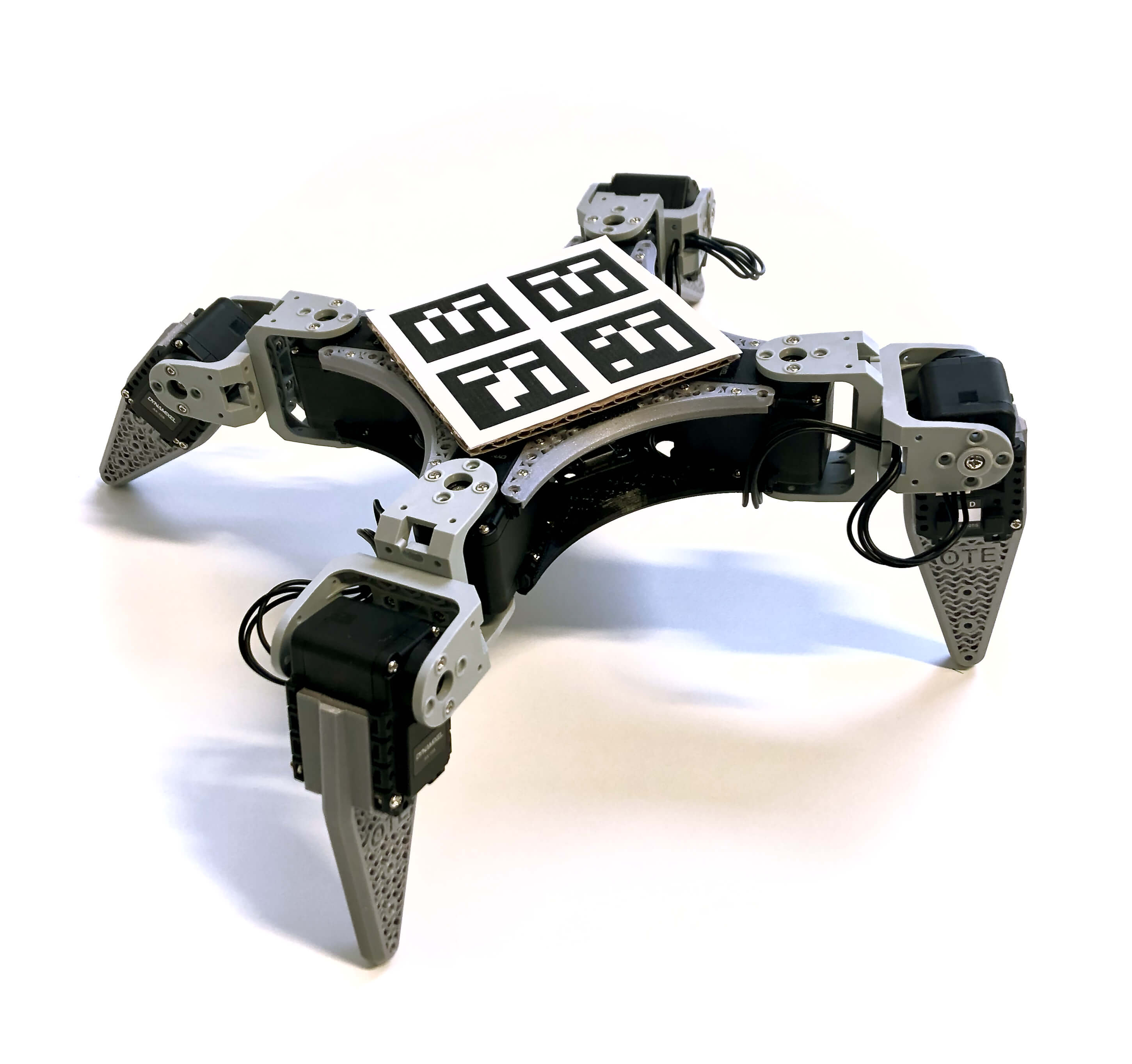}
      };}] at (0,0) {};
  \end{tikzpicture}
  \caption{The low-cost RealAnt robot imitating the `Ant' RL benchmark.}
  \label{fig:photo}
\end{figure}

\begin{table}[t]
\vspace{4mm}
\caption{Bill of materials grouped into core robot parts, tracking equipment, and a wireless extension package.}
\label{t:cost}
\centering
\setlength{\tabcolsep}{4pt}
\begin{tabularx}{\columnwidth}{llccc}
\toprule
& &  & \sc Unit & \sc Total  \\
& \sc Component & \sc QTY & \sc Price (\euro{}) & \sc Price (\euro{}) \\
\midrule
\parbox[t]{2mm}{\multirow{6}{*}{\rotatebox[origin=c]{90}{Robot}}}
& Dynamixel AX-12A servos & 8 & 40 & 320 \\

& OpenCM9.04-A microcontroller & 1 & 10 & 10 \\

& OpenCM9.04 accessory set & 1 & 6 & 6 \\

& 3D printed parts & --- & --- & 3 \\

& Screws and cables & --- & --- & 15 \\

\cmidrule{2-5}

& Total & & & 354 \\

\midrule

\parbox[t]{2mm}{\multirow{2}{*}{\rotatebox[origin=c]{90}{Track}}}

& Web cam (\eg, Logitech Brio 4K) & 1 & 250 & 250 \\

& Printed tags on office paper & 1 & 1 & 1 \\

\midrule

\parbox[t]{2mm}{\multirow{4}{*}{\rotatebox[origin=c]{90}{Wireless}}}

& Bluetooth-serial converter HC-06 & 1 & 9 & 9 \\

& LiPo battery 4S 3.3Ah & 1 & 30 & 30 \\

& LM2596 12V buck converter & 1 & 2 & 2 \\

& Low voltage buzzer & 1 & 3 & 3 \\




\bottomrule
\end{tabularx}
\end{table}







The contributions of this paper are as follows.
{\em (i)}~We develop a low-cost minimal quadruped robot called RealAnt, a physical version of the popular Ant benchmark used in reinforcement learning research.
{\em (ii)}~We develop the supporting software stack to perform RL on the physical platform. We also provide simulated versions of the RealAnt robot (with same dimensions and state-action spaces, and delayed noisy observations) in PyBullet and MuJoCo simulators for rapid testing.
{\em (iii)}~We validate that the robot is suitable for real-world RL research, propose three benchmark tasks, and report baseline results on these tasks using REDQ, TD3 and SAC algorithms. We evaluate the state-of-the-art REDQ reinforcement learning algorithm for the first time on a physical platform.
Hardware design, supporting software, RL baselines, and a video of learned gaits are available at: \url{https://github.com/AaltoVision/realant-rl/}.

\section{Related Work}
Scalable low-cost robot platforms can enable a plethora of real-world applications like last-mile delivery and automate highly repetitive manually laborious tasks like object stacking. The robotics community is actively working towards more affordable robots, and in this section, we review recent works on low-cost platforms for robotics research.

Recent works have proposed designs for affordable quadruped robots.
The MIT mini-cheetah \cite{katz2019mini}, costing around \$10k, is a small quadruped robot that can perform a wide range of locomotion behaviors.
Solo \cite{grimminger2020open}, costing around 4k~\euro{}, is an open-source, lightweight, and torque-controlled quadruped robot based on low complexity actuator modules using brushless motors.
Stanford Doggo \cite{kau2019stanford}, costing less than 3k~\euro{}, is an open-source quadruped robot based on a quasi-direct-drive mechanism. While these robots are designed for motion planning controllers, we propose RealAnt, costing less than 500~\euro{}, with a focus on research in real-world reinforcement learning. Unlike other available quadruped robots, RealAnt is designed as a direct analogy of the popular Ant benchmark.

Similar to RealAnt, ROBEL \cite{ahn2020robel} is a recently introduced open-source platform for benchmarking real-world reinforcement learning. The ROBEL platforms consist of two robots: D'Claw and D'Kitty, for manipulation and locomotion tasks respectively. D'Kitty is a a 12 DoF quadruped robot, costing around \$4.2k, with Dynamixel XM430-W210-R smart actuators.
RealAnt is significantly cheaper as it is an 8~DoF robot with cheaper Dynamixel AX-12A servos.
While D'Kitty relies on a HTC Vive Tracker setup (costing more than 500~\euro{}) for pose estimation, we use simple fiducial marker tracking (only requiring a web camera) for the same. Being so cheap, it is possible to build more than ten RealAnt robots at the cost of a D'Kitty, enabling scalable and broader real-world experiments. While only simulated or sim2real experiments have been demonstrated using D'Kitty \cite{ahn2020robel}, we demonstrate data-efficient RL from scratch directly on the RealAnt robot.

While there exist several cheap quadruped robots, mostly proposed for educational purposes, they were not designed to sustain the abuse of reinforcement learning. 
Reinforcement learning involves aggressive exploratory actions that can easily damage the servos or the 3D printed body of the robot. We validate that the proposed RealAnt robot can sustain such aggressive actions.

Wheeled robots tend to be more affordable. Examples of such research platforms include Donkey car, DeepRacer~\cite{balaji2019deepracer}, JetBot, DuckieBot~\cite{paull2017duckietown} and their costs fall in the range of \$250 to \$500.
RealAnt enables research and education on contact-rich legged locomotion in a similar affordable cost range.

There has also been progress in reducing the cost of robot platforms for manipulation tasks. Recent works \cite{gupta2018robot, gealy2019quasi} have proposed such platforms in the cost range of \$2k to \$5k. REPLAB \cite{yang2019replab}, costing around \$2k, is an easily reproducible benchmark for vision-based manipulation tasks.

Compared to existing solutions, RealAnt a direct analogy of the popular Ant benchmark, and is hence well suited for bridging the gap to real-world applications of RL.

\section{RealAnt}

\begin{figure}[t]
\centering
\vspace{2mm}
\includegraphics[width=\columnwidth, trim=20 100 40 60, clip]{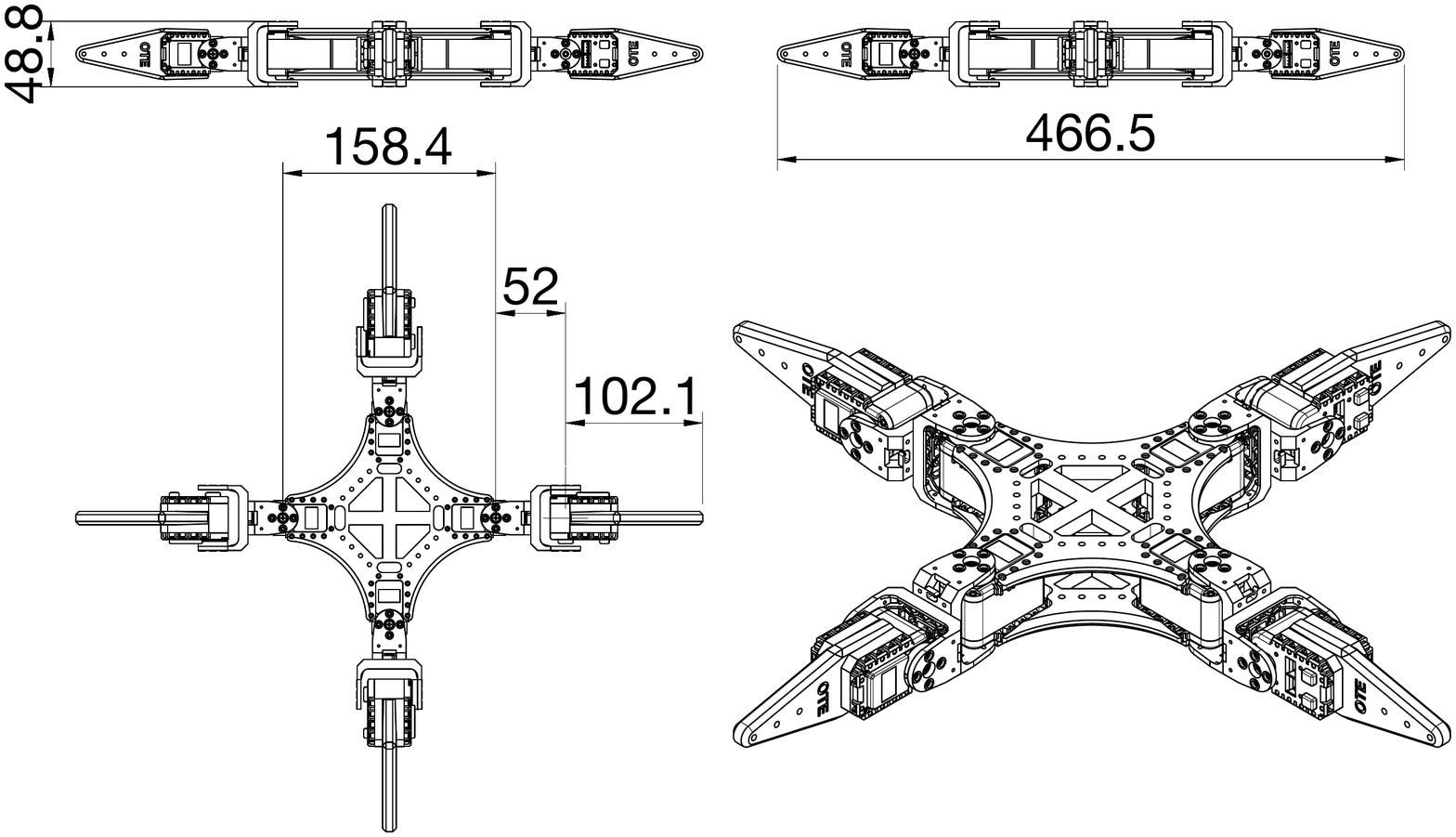}
\caption{Schematic details of the RealAnt robot (all units in millimeters).}
\label{fig:hardware}
\end{figure}

We design RealAnt, a minimal and low-cost physical version of the Ant benchmark for research in real-world reinforcement learning. Similar to the Ant benchmark, RealAnt is an 8 DoF quadruped robot (see Fig.~\ref{fig:photo} for a photo). RealAnt is based on easily available electronic components and a 3D printed body. 
List of all components used in RealAnt and their costs are in Table~\ref{t:cost}.
The RealAnt can be assembled from these components in less than an hour, by using a Phillips screwdriver, side cutters, and a soldering iron.

\subsection{Mechanical Design}

The minimally designed body of the robot consists of 1)~four 3D printed legs, 2) eight Dynamixel AX-12A servos (and eight FP04-F2 frames sold with them), and 3) a 3D printed top and bottom torso.
Each leg of the robot constitutes of two Dynamixel servos joints affixed to each other using Robotis FP04-F2 frames. 
Four of the leg assemblies are joined together using a 3D printed torso top and bottom plates. 3D printers are easily accessible and allow for rapid prototyping and cost-effective manufacturing.
The schematic details are illustrated in Fig.~\ref{fig:hardware}.

The parts were printed in PLA (Prusament filament) using a consumer 3D printer (Creality Ender 3 v2). A complete set of parts requires 
13.5~h to print, for two torso plates and four legs. To lower the printing time and produce rigid enough parts, they were printed using 0.2~mm layer height, 20\% gyroid \cite{schoen1970infinite} infill, and with open top and bottom layers. The printed parts weigh 106~g in total, 
 costing 2.65~\euro{} in filament costs, assuming a filament price of 25~\euro{}/kg. The total robot weight is around 710~g. 
The parts are designed to be long-lasting, but if they break, they can be easily reprinted locally and since PLA is biodegradable, they can be recycled.

Economical servo motors are challenging to use in an RL setting. The random actuation and hard hits to the floor can wear and break down the small gears in servo gearboxes. Also in long continuous operation under load, some servos tend to overheat and break. To overcome these issues, we designed the legs short and the platform lightweight enough to reduce sharp jerks, and we opted to do 10-second episodes. Between episodes, the robot position is reset manually if necessary. Also, in software, we limited the maximum torque of the servo motors to half. 

Initially, we used high-torque RC hobby servos (such as Turnigy TS-910) for trials, but Robotis Dynamixels were eventually selected for the design due to their longevity in testing, owing mostly to adjustable torque limits, in-built temperature sensors, and overall build quality.

\subsection{Electrical Design}

\begin{figure}[t]
\centering
\vspace{2mm}
\includegraphics[width=\columnwidth, trim=60 95 380 305, clip]{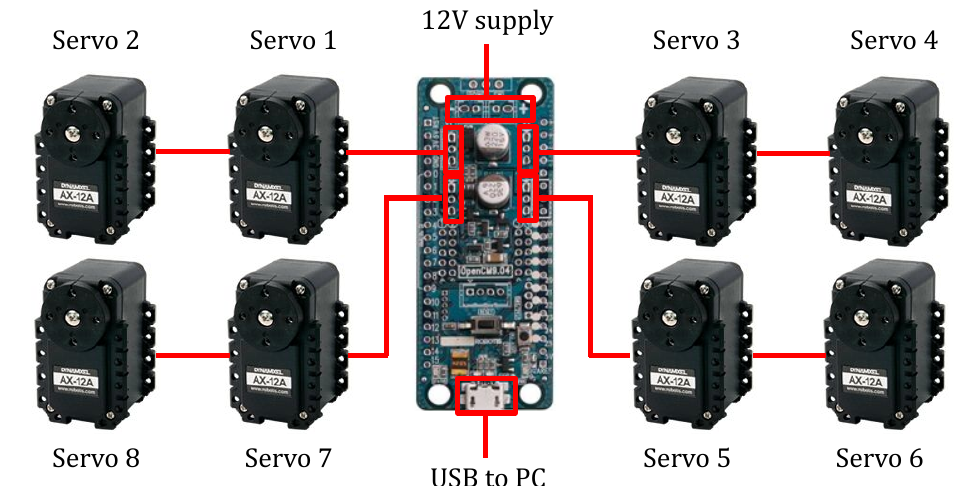}
\caption{Electrical connectivity of the RealAnt robot.}
\label{fig:electrical}
\end{figure}

See Fig.~\ref{fig:electrical} for an illustration of the electrical connectivity of the RealAnt robot.
For a simple and reliable experimental setup, we use an external lab power supply and directly control the robot using a wired USB connection from a computer. Both the USB and power wires are connected to the OpenCM9.04 microcontroller board. The leg servos are daisy-chained and each leg is connected to one of the four 3-pin servo ports on the board.
Alternatively, a LiPo battery, a buck DC-DC voltage converter, a low battery voltage buzzer and a Bluetooth serial converter can be used for completely wireless operation without using wired power and data cables. 

\subsection{Pose Estimation}

The state of the RealAnt robot includes the 6 DoF pose of the robot and this information is also necessary to derive reward functions for reinforcement learning. For example, the reward used in the simulated Ant benchmark is the forward velocity of the robot.
We rely on augmented reality (AR) tag tracking using ArUco tags and the OpenCV version of the popular ArUco library \cite{garrido2016generation, romero2018speeded,bradski2000opencv} for pose estimation by detection of square fiducial markers. The tags are printed on A4 office paper and glued to cardboard for rigidity. We attach the tracking tag to the top of the robot body. We use a Logitech Brio 4K web camera and place a frame reference tag within the camera view. The camera model is calibrated by taking pictures of a chessboard pattern.

Using a consumer web camera for the position estimation can be challenging, due to camera latency and frame timing jitter. The latency can be as long as several hundred milliseconds, and frame timing jitter considerable. Using the Logitech Brio 4K web camera with $1280{\times}720$ resolution at 60~fps, we measured latency of around 110~ms. This latency requires a robust RL approach and is accordingly added and tested in the simulation model.

Tag-based pose estimation is noisy
and since velocity estimation is important in our tasks as a reward signal (see Sec.~\ref{sec:tasks}), we used Holoborodko's smooth noise-robust differentiator \cite{holoborodko2008smooth} to improve the velocity estimates. Furthermore, as the web camera is positioned on top of the tags, the $z$ (depth) axis measurement is very noisy. We additionally smoothed this with a lowpass filter. We also add and test this estimation noise in our simulation model.

\subsection{Software Design}


\begin{figure}[t]
\centering
\vspace{2mm}
\includegraphics[width=\columnwidth, trim=40 338 265 80, clip]{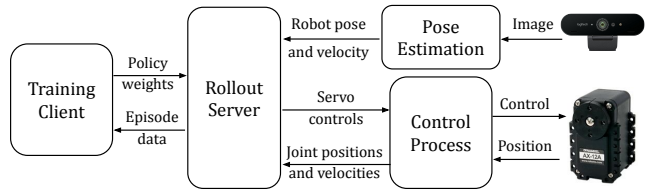}
\caption{Software design of the RealAnt robot.}
\label{fig:software}
\end{figure}

We provide supporting software for the RealAnt platform so that it can be easily used and our experiments easily reproduced.
We decouple the software into four processes that communicate using ZeroMQ (much like a lightweight ROS environment): a training client, a rollout server, a pose estimation process, and a control process. See Fig.~\ref{fig:software} for an illustration of how these processes communicate.
The training client controls the whole learning process. It sends the latest policy weights to the rollout server to initiate a training episode. The rollout server loads the policy weights, collects the latest observations from the control process and the pose estimation process, and sends the action computed using the policy network back to the control process. The control process continuously collects servo measurements from the microcontroller and publishes them to the rollout server. It also subscribes to actions from the rollout server and applies them to the robot through the microcontroller. The pose estimation process continuously collects images from web camera, computes pose estimates and publishes them to the rollout server. After completing an episode, the rollout server sends back the collected episode data to the train client. The newly collected data is added to a replay buffer and the agent is updated a few times by sampling from this replay buffer. Decoupling of these processes allow them to run seamlessly in different machines. For example, the data collection (with rollout server and control process) can be performed on a low-end computer and training can be performed on a high-end computer.

The rollouts were performed on a ThinkPad L390 Yoga laptop, which was attached to the RealAnt board and to the webcam. Training was done on a Linux desktop machine equipped with an Nvidia GTX1080 GPU.
One episode cycle of REDQ took approximately 110~s wall clock time, including rollout and training. 


Based on our hardware design, we provide simulated versions of the robot on PyBullet and MuJoCo simulators, for rapid testing and development.
The simulated robot has the same physical dimensions, state-action spaces, and roughly the same dynamics. We find that MuJoCo simulation is more stable and easy to modify while the PyBullet simulator is free and easily accessible. We run our experiments on both simulators for ease of reproducibility.

\begin{figure*}[t]
\centering
\vspace{2mm}
\includegraphics[width=\linewidth, trim=10 10 10 10, clip]{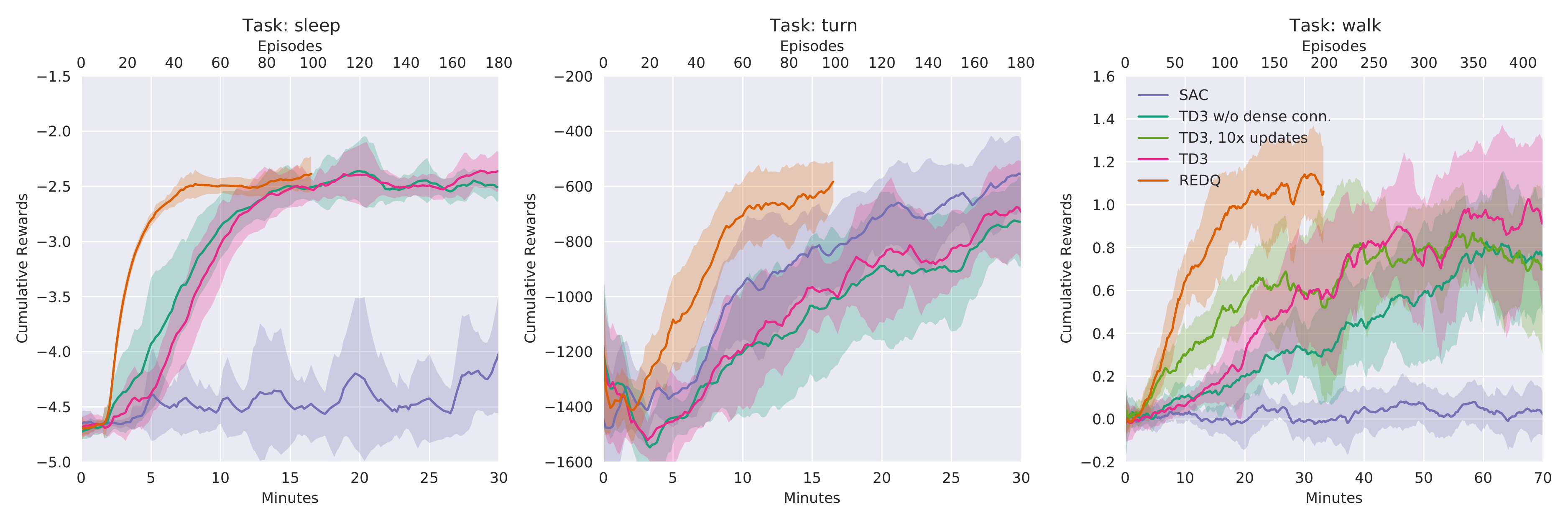}\\[-6pt]
\caption{Results on the MuJoCo simulator: SAC and TD3 learning algorithms, with and without dense connections. We plot the mean and std of the cumulative rewards of 10 runs for each setting. TD3 performs better than SAC and dense connections lead to better sample-efficiency in the walk task.}
\label{fig:mujoco-results}
\end{figure*}

\section{Benchmark Tasks}
\label{sec:tasks}

\begin{figure*}[t]
\centering
\vspace{1mm}
\includegraphics[width=\linewidth, trim=10 14 10 10, clip]{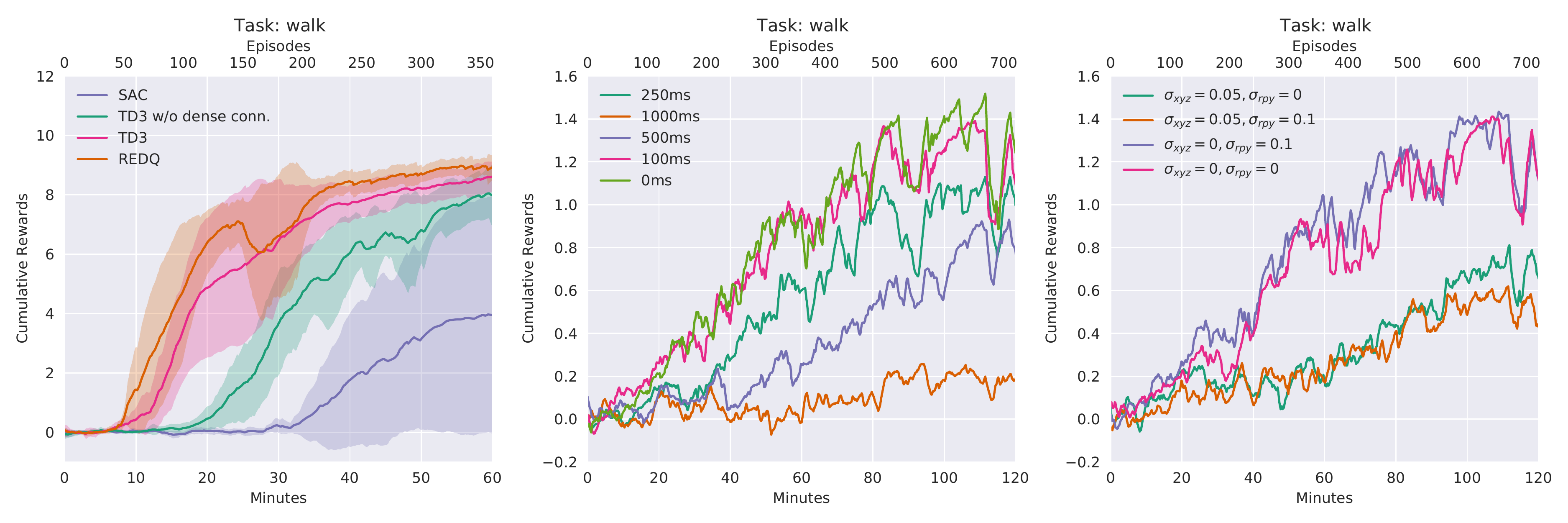}\\[-6pt]
\caption{
\textbf{Left}: Results on the PyBullet simulator: Similar to MuJoCo, we observe that TD3 performs better than SAC and dense connections improve the sample-efficiency of TD3.
\textbf{Middle}: Results of our ablation study on the effect of observation latency. Observation stacking is able to effectively handle latencies of up to 100~ms without any impact on performance. Larger latencies degrade learning efficiency. 
\textbf{Right}: Results of our ablation study on the effect of tracking noise. We add zero-mean Gaussian noise to body $xyz$ and $rpy$ observations with standard deviations of $\sigma_{xyz}$ and $\sigma_{rpy}$, respectively. Observation stacking is able to deal with noisy observations but noisy rewards (computed from noisy $xyz$ estimates) clearly degrades learning efficiency.
}
\label{fig:ablation-results}
\end{figure*}

In this section, we define the RL specifics of the robot including the observation and action spaces. We then introduce three benchmark tasks used to evaluate RL algorithms on simulators and the physical platform.

We use an episodic formulation of reinforcement learning. Each training session is split into episodes lasting 200 steps. We manually reset the robot to the starting position of the task before each episode. We use a control frequency of 20Hz so each episode corresponds to 10 seconds of experience.

The 8-dimensional action space of the RealAnt robot defines the set-point for the angular position of the robot joints.
The observation space of RealAnt is computed from 6D pose estimates and joint positions. The 3D positions ($x$, $y$, and $z$) and 3D angles (roll $\alpha$, pitch $\beta$, and yaw $\gamma$) of the torso are obtained from pose estimation using AR tag tracking. The torso velocities are computed using differences of consecutive pose estimates. The angular positions and velocities of the joints are obtained from the joint servos. The 29-dimensional observation space of the RealAnt robot consists of:
\begin{enumerate}
    \item $x$, $y$, and $z$ velocities of the torso (3),
    \item $z$ position of the torso (1),
    \item $\sin$ and $\cos$ values of Euler angles of the torso (6),
    \item velocities of Euler angles of the torso (3),
    \item angular positions of the joints (8), and
    \item angular velocities of the joints (8).
\end{enumerate}

The three benchmark tasks introduced below share the same observation and action spaces. Only the rewards $R_t$ are different.

\paragraph{Stand / Sleep}
This is a very simple task that involves attaining a goal torso height ($z_g$). That is, 
\[ R_t = -\|z_t - z_g\|^2, \]
where $z_t$ is the $z$ position of the robot torso at timestep $t$. In the simulator, the robot starts each episode standing upright and the goal is to \emph{sleep} (that is, $z_g = 0$) and in the physical experiments, the robot starts each episode lying down and the goal is to \emph{stand} upright (that is, $z_g = 0.12$).

\paragraph{Turn}
This task involves the robot turning 180\degree{} to face the other direction. Each episode starts with the robot lying down. The robot has to balance itself and coordinate all joints to turn the whole body around. The initial yaw $\gamma_0$ of the robot is 0 and the goal in this task is to rotate to a yaw of 180\degree{} or 3.14 radians. The rewards are computed based on this angular distance:
\[ R_t = -\|\gamma_t - 3.14\|^2. \]
This challenging task can also be used to ensure that pose estimation and tracking is accurate.

\paragraph{Walk}
This task is the same as for the original Ant benchmark used in simulated RL experiments: learning to walk forward as fast as possible. Each episode starts with the robot lying down. This challenging task involves the robot coordinating all its joints to walk forward as fast as possible. The reward in this task is the forward velocity of the robot:
\[ R_t = \dot{x}_t, \]
where $\dot{x}_t$ is the velocity of the robot along the $x$ axis.

\section{Experiments}

We begin each training session with ten episodes of data collected using a random policy. We alternate between data collection and training for every episode. 
We use the state-of-the-art REDQ \cite{chen2021randomized} algorithm and the popular TD3 \cite{fujimoto2018addressing} and soft actor-critic (SAC) \cite{haarnoja2018softa, haarnoja2018softb} algorithms in this paper. REDQ allows for data-efficient RL due to high number of gradient updates of a randomized ensemble of 10 Q-networks. We perform 2000 learning updates of REDQ or 200 updates of SAC/TD3 at the end of each episode. We use the Adam optimizer with learning rate 0.0003 to update all parameters.

We use fully-connected networks with dense connections for our policy and value networks. As reported in \cite{sinha2020d2rl}, we find that dense connections (concatenation of network inputs to the input of each layer) enable stable training of deeper networks, which allows for improved sample-efficiency. We use dense connections, three hidden layers, and 256 hidden units for our policy and value networks.


Reinforcement learning on the physical robot has to inevitably deal with latencies and noise in the observations. This makes the environment non-Markovian. To deal with this, we construct the robot state (to be used by the RL algorithm) by stacking the past four observations. Note that noise in the observations also leads to noisy rewards, which makes learning even more challenging. We introduce Gaussian noise with standard deviation 0.01 and an observation latency of 2 steps (100ms) into the simulator environments to better match real-world conditions.

\subsection{Results on Simulator}


We first test the state-of-the-art REDQ, SAC and TD3 algorithms on the simulated versions of our RealAnt robot. The results of our experiments on the MuJoCo simulator are shown in Fig.~\ref{fig:mujoco-results}. We observe that REDQ significantly outperforms TD3 and SAC on all tasks, to learn them in less than 15 minutes of experience. TD3 is also able to learn all tasks but requires more data. SAC is able to perform slightly better than TD3 in the turn task but performs poorly in sleep and walk tasks. We also perform ablation studies on the walk task to study the importance of dense connections used by all algorithms and the high number of learning updates used by REDQ. The use of dense connections lead to significantly better learning performance. TD3 with 10x learning updates (matching the REDQ algorithm) learns faster in the first 150 episodes but then its performance deteriorates.

We also test the RL algorithms on the PyBullet simulator. We find that RL algorithms are able to exploit the instabilities of the PyBullet simulator, so we only test on the walk task. The results of our experiments are shown in Fig.~\ref{fig:ablation-results}. Similar to our observations in experiments on the MuJoCo simulator, we find that REDQ outperforms TD3 and both algorithms significantly outperform SAC.

\begin{figure*}[t]
\centering
\vspace{1mm}
\includegraphics[width=\linewidth, trim=10 13 10 10, clip]{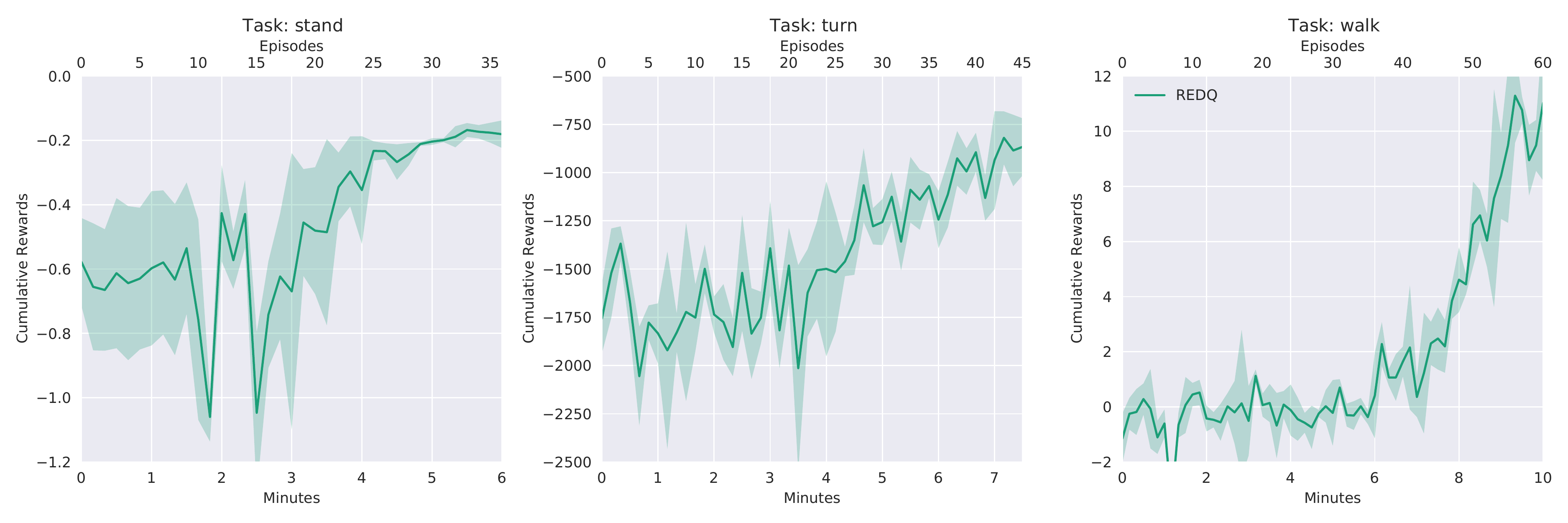}\\[-6pt]
\caption{
RealAnt robot can learn all tasks (using the REDQ algorithm) within 10 minutes of data. We focus on data efficiency and terminate the experiment once the policy  performs consistently well for 10 consecutive episodes. In the walk task, the robot has learned to successfully walk outside the tracking area. We plot the mean and std of three runs on each task.}
\label{fig:real-results}
\end{figure*}

\begin{figure*}[t]
\centering
\includegraphics[width=\linewidth, trim=0 0 0 0, clip]{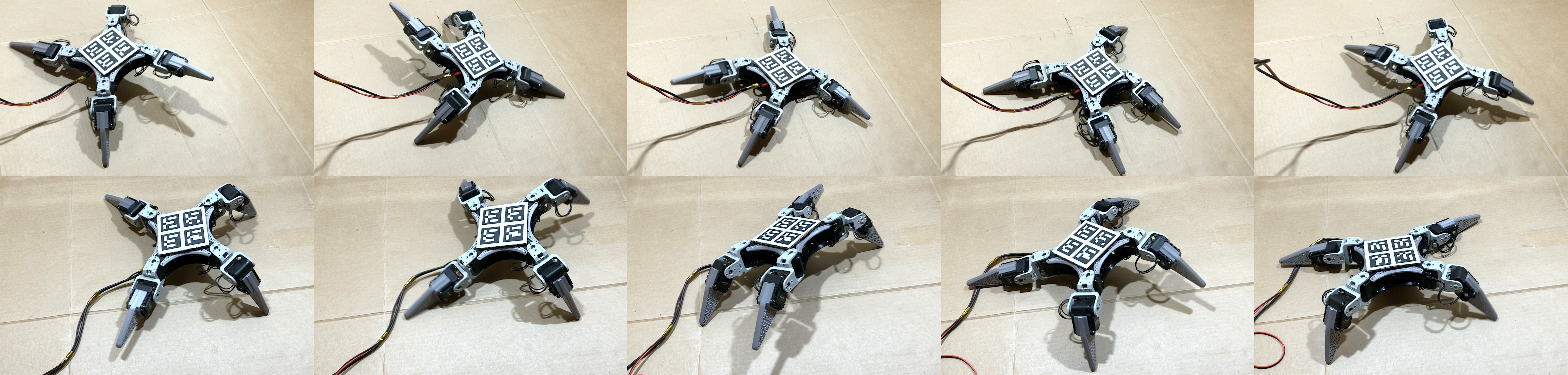}\\[-6pt]
\caption{Example of learned walking (top row) and turning (bottom row) gaits after 10 minutes of training data on the RealAnt robot.}
\label{fig:real-gait}
\end{figure*}

\subsection{Ablation Studies on Simulator}

Real-world experiments inevitably consist of delays and noise. Inexpensive USB-connected web cameras often have relatively high communication bus and processing latency of several hundred milliseconds. Furthermore, the pose estimation and tracking is usually noisy. To ensure our method works on the real robot, we study the effect of such delays and noise on the learning efficiency of TD3 (since it is similar to REDQ, runs significantly faster in wall-clock time, and can robustly learn all tasks). We use the MuJoCo simulator for these ablation studies.

\subsubsection{Effect of tracking latency}

We test the effect of tracking latency by delaying the observation of body position and orientation values in the simulator. To deal with this latency, we simply stack observations from past steps (larger than latency). The results of our experiments are shown in Fig.~\ref{fig:ablation-results}. We find that observation stacking is able to effectively deal with even significant delays of 10 steps or 500~ms.

\subsubsection{Effect of tracking noise}

We test the effect of tracking noise by adding different levels of noise to observations of body position and orientation values. The results of our experiments are shown in Fig.~\ref{fig:ablation-results}. We find that observation stacking used to deal with latency is also effective in dealing with observation noise. The learning algorithm is sensitive to measurements such as body position which is used to derive rewards. However, it is very robust to the other measurements such as body orientation.

\subsection{Results on Physical Robot}

\begin{table}[t]
\caption{Walking speed of RealAnt on different surfaces using REDQ policies pretrained on a vinyl flooring}
\label{t:surface}
\centering
\begin{tabularx}{0.96\columnwidth}{ccccc}
\toprule
Vinyl & Hardwood & Polyester Rug & Cardboard & Plastic \\
\midrule
5.96 cm/s & 4.27 cm/s & 6.51 cm/s & 6.02 cm/s & 5.83 cm/s \\
\bottomrule
\end{tabularx}
\end{table}

In this section, we validate the RealAnt robot by evaluating the best performing REDQ algorithm on the three proposed benchmark tasks of stand, turn, and walk. We use the same network architecture (with dense connections) and hyperparameters as in our simulator experiments. 

We focus on learning a reasonable policy in a data-efficient manner and terminate the training once the policy performs consistently well (based on visual observation) for more than 10 subsequent episodes. Between episodes, the robot position is optionally reset manually so as to maintain accurate pose estimation throughout the training episode. The results of our experiments are shown in Fig.~\ref{fig:real-results}. We are able to successfully learn all three tasks from just 60 episodes or 10~min of real-world experience, which demonstrates that training multiple different tasks with the RealAnt physical robot is feasible within a single work-day. An example of a learned walking gait and a turn is shown in Fig.~\ref{fig:real-gait}. In the walk task, while it is possible to learn better gaits with longer training, the policy learns to successfully and consistently walk outside the field of view of our AR tag tracking setup. Implementing other challenging tasks that can be performed within the field of view or use of better tracking systems with a larger tracking area is a line of future work.

To test the robustness of our learned walk policy, we test it's performance of different surfaces. The training was performed using REDQ algorithm on vinyl flooring and tested on a hardwood floor, a polyester rug, a flat cardboard, and an IKEA KOLON plastic floor protector. We measure the walking speed of the robot on these different surfaces and report them in Table~\ref{t:surface}. The results are averaged across three pre-trained policies evaluated for 5 episodes each. The robot trained on a vinyl flooring walks at a similar speed on a flat cardboard and a plastic floor protector. It walks even faster on the polyester rug as it provides more friction. It walks slower on a hardwood floor as it is slippery. While the walking speed varies according to the friction of the surfaces, the robot is able to robustly walk forward in all of them.



\section{Discussion}



\subsection{Pose Estimation}


Getting the AR tag-based pose estimation to work well with an aggressively moving robot is a challenge due to motion blur. This can make the AR library incapable of detecting the AR tag and consequently providing a pose estimation. The exposure time on web camera needs to be short enough to overcome the motion blur. Therefore, the tracking performance was tested before attempting robot learning, by adding enough lighting and by tweaking camera exposure settings. Some LED floodlights can also produce flickering light (similar to fluorescent lights), which becomes evident only when using short exposure times. This can affect the AR tag detection performance, but slight flicker was not an issue for our experiments.
The AR tag detection 
is also sensitive to camera angle and height adjustment. Some trial and error is required to find a sweet spot where it works well. This ambiguity possibly could be removed with 3D-printed structure under the AR tags (e.g. a shallow pyramid).

Learning to control the robot based on feedback from tag-based pose estimation
can be challenging. Especially, for the walking task, the reward only consists of the forward velocity. Estimating the forward velocity from noisy position measurements by naive numerical differentiation leads to even more noisy velocity estimates, which greatly affected the learning performance in our experiments. Therefore, we use Holoborodko’s smooth noise-robust differentiator \cite{holoborodko2008smooth} to obtain better velocity estimates from real-world sensor data, which smoothed the estimates enough to learn efficiently.

\subsection{Robot Design}

During tests of several iterations of the robot design, several hardware design considerations became evident. During random exploration and training, the robot and its joints get a harsh beating. Hard impacts to the floor can break 3D-printed rigid structures and strip the servo motor gearbox gears. Also, continuous random actions can destroy servo motor electronics by overheating.

For joint actuation, we initially used RC servos (such as Turnigy TS-910). The problem with general-purpose RC servos is that they don't have easily controllable torque limits and no overheat protection. Thus, we opted for Dynamixel smart actuators that have torque (current) limits so that hard impacts and jerks are reduced. The Dynamixels have also an embedded temperature sensor that is used for motor and electronics overheat protection. We use Dynamixel AX-12A servos in our robot but the newer and similarly priced Dynamixel XL430-W250-T is a good alternative, even though the AX-12A is cheaper in bulk pricing.

We designed the the 3D-printed parts to be lightweight and slightly elastic. Especially, the legs and body servo mounts have a little bit of flex in them, to reduce impacts. To obtain lightness and elasticity, we utilized a gyroid infill pattern which likely spreads stresses more evenly across the parts. We also printed the parts without top and bottom layers, which also helped to make it more lightweight and more elastic. We experimented with PETG and PLA filaments, and chose PLA because it has better availability, it is easier to print and it is also biodegradable.

\subsection{Use in Education}

A quadruped robot learning to walk by itself could inspire curiosity in students to learn about programming, robotics, and artificial intelligence. Building and programming the RealAnt robot involves simple yet interesting steps that could serve as an introduction to robotics and programming. For example, a curriculum designed around RealAnt could involve introductions to: programming a microcontroller, Python programming language, controlling servo motors, 3D modelling, 3D printing, augmented reality, and machine learning. Such a curriculum could equip students with the knowledge to design and build similar robots by themselves. Since RealAnt was designed to sustain aggressive explorations in RL, students can explore servo controls without fear of breaking parts. Any broken part can be easily and cheaply 3D printed and replaced.


\section{Conclusions and Future Work}


We introduce a very low-cost and minimal robot platform called RealAnt, for real-world educational and research use in reinforcement learning of legged locomotion. RealAnt is based on the Ant benchmark that is very familiar to researchers in RL, allowing for straightforward testing of existing models and algorithms. 
We provide the supporting software to perform RL research on RealAnt.
We validate the robot with RL experiments on three benchmark tasks and demonstrate successful learning using REDQ and TD3 algorithms. Model-based RL algorithms have demonstrated significantly better sample-efficiency on the simulated Ant benchmark \cite{janner2019trust, boney2020regularizing}. However, such algorithms also require significantly more computational resources, and testing them on RealAnt is a line of future work.

\addtolength{\textheight}{-4cm}   


We found that off-the-shelf tracking cameras such as Intel RealSense T265 are highly sensitive to the high-frequency vibrations in legged robots like RealAnt, leading to significant drift in tracking. Visual-inertial odometry algorithms that are robust to such vibrations can be used for pose estimation and tracking without any external AR tag tracking or motion capture systems.
While reinforcement learning could damage the components of a robot, the learning does not take this into account. Feedback from energy usage, foot contact sensors, servo temperature, etc. can be used for safety-aware learning. Sparse negative feedback can also be provided when RL causes significant damage to the robot such as breaking of 3D printed parts or servo gears. Safe exploration in RL is an active research area.






\section*{Acknowledgment}

The authors thank Risto Bruun (Ote Robotics) and Filip Granö for their contributions to the hardware and software design along the development of the RealAnt, Sagar Vaze (University of Oxford) for his suggestions with the manuscript and also Curious AI founders Harri Valpola, Antti Rasmus, Mathias Berglund, and Timo Haanpää for support and inspiring early model-based experiments with the simulated Ant.


\bibliographystyle{IEEEtran}
\bibliography{IEEEabrv,root}

\end{document}